\documentclass{article} % For LaTeX2e
\usepackage{arxiv,times}

\usepackage{amsmath,amsfonts,bm}

\def\eqref#1{equation~\ref{#1}}
\def\1{\bm{1}}

\DeclareMathAlphabet{\mathsfit}{\encodingdefault}{\sfdefault}{m}{sl}
\SetMathAlphabet{\mathsfit}{bold}{\encodingdefault}{\sfdefault}{bx}{n}

\DeclareMathOperator*{\argmax}{arg\,max}

\usepackage{hyperref}
\usepackage{url}
\hypersetup{
	colorlinks=true,
        breaklinks=true,
        citecolor={blue!50!black},
}
\usepackage{appendix}
\usepackage{subcaption}
\usepackage{multirow}
\usepackage{graphicx}
\usepackage{color}
\usepackage{booktabs}
\usepackage{array}
\usepackage[labelfont=bf]{caption}
\usepackage{makecell}
\usepackage{xcolor}
\usepackage{diagbox}
\usepackage[T1]{fontenc}
\usepackage[capitalize,noabbrev]{cleveref}

\makeatletter
\def\@fnsymbol#1{\ensuremath{\ifcase#1\or \dagger\or \ddagger\or
   \mathsection\or \mathparagraph\or \|\or **\or \dagger\dagger
   \or \ddagger\ddagger \else\@ctrerr\fi}}
\makeatother

\title{
Efficient Residual Learning with Mixture-of-Experts for Universal Dexterous Grasping
}

\author{%
  Ziye Huang$^1$, Haoqi Yuan$^1$, Yuhui Fu$^1$, Zongqing Lu$^{1,2}$\thanks{Correspondence to Zongqing Lu <zongqing.lu@pku.edu.cn>.} \\  \\
   $^1$Peking University \\
   $^2$Beijing Academy of Artificial Intelligence
}

\iclrfinalcopy % Uncomment for camera-ready version, but NOT for submission.
\begin{document}

\maketitle

\begin{abstract}
Universal dexterous grasping across diverse objects presents a fundamental yet formidable challenge in robot learning. Existing approaches using reinforcement learning (RL) to develop policies on extensive object datasets face critical limitations, including complex curriculum design for multi-task learning and limited generalization to unseen objects. 
To overcome these challenges, we introduce ResDex, a novel approach that integrates residual policy learning with a mixture-of-experts (MoE) framework. ResDex is distinguished by its use of geometry-unaware base policies that are efficiently acquired on individual objects and capable of generalizing across a wide range of unseen objects. Our MoE framework incorporates several base policies to facilitate diverse grasping styles suitable for various objects. By learning residual actions alongside weights that combine these base policies, ResDex enables efficient multi-task RL for universal dexterous grasping.
ResDex achieves state-of-the-art performance on the DexGraspNet dataset comprising 3,200 objects with an 88.8\% success rate. It exhibits no generalization gap with unseen objects and demonstrates superior training efficiency, mastering all tasks within only 12 hours on a single GPU.
\end{abstract}

\section{Introduction}

Dexterous robotic hands \citep{multifingered-review, leaphand} provide advanced capabilities for complex grasping tasks, similar to those performed by human hands. However, achieving universal dexterous grasping across a wide range of objects remains a significant challenge due to the high degrees of freedom (DoFs) for dexterous hands and the high variability in object geometry in the real world. Previous works \citep{dexpoint, agarwal2023dexterous} develop dexterous grasping policies using reinforcement learning (RL), but these policies are limited to a small range of objects that are similar to the training objects. To improve the scalability of universal dexterous grasping, recent studies \citep{dexycb, wang2023dexgraspnet, hang2024dexfuncgrasp} introduce datasets that contain a wide variety of objects, each labeled with grasping poses. \citet{xu2023unidexgrasp, wan2023unidexgrasp++, wu2024unidexfpm} leveraged these datasets to learn universal grasping policies through a teacher-student framework, which addresses the challenges of multi-task optimization. They first train state-based policies using RL to master all objects within the dataset, and then distill these policies into a universal vision-based policy.

However, these approaches exhibit certain limitations. UniDexGrasp \citep{xu2023unidexgrasp} involves a complicated curriculum learning design, requiring iterative training across an expanding set of objects, which significantly increases training time and necessitates careful design for the curriculum.  Similarly, UniDexGrasp++ \citep{wan2023unidexgrasp++} requires training various state-based policies on a large number of object clusters. This not only consumes substantial training time but may also lead to overfitting, as training is conducted individually on separate object groups.
In this study, we investigate {\textit{how to directly learn a multi-task dexterous grasping policy across thousands of objects}}, which enables both efficient learning and enhanced generalization.

% \red{P3: Technical motivation and method: (1) (dexterous functional grasping) motivates us to train a base policy using less observations. Though have a lower performance, it exhibits more robust generalization across objects  (2) residual RL (xxx) provides an efficient way to improve policy given a base policy. (3) We propose to learn a geometry-unaware policy generalizable to many objects, then learn a residual policy using multi-task RL on all objects. }

Residual policy learning \citep{silver2018residual, johannink2019residual} offers an efficient approach to learning challenging tasks by training a policy to output residual actions using RL, where a suboptimal base policy is provided. This approach has the potential to address the optimization challenges in multi-task RL \citep{wu2024learning}, particularly when the base policy can effectively explore all tasks. Motivated by this, we propose \textbf{ResDex} to train a residual multi-task policy for universal dexterous grasping. The key question then becomes, {\textit{how to efficiently acquire a base policy that possesses some generalizability to grasp a wide range of objects?}} Directly applying multi-task RL to all objects leads to worse results due to multi-task gradient interference \citep{gradient-surgery} and requires extensive training time. Conversely, training a policy to grasp a specific object often results in poor generalization to unseen objects. 

Recent work \citep{agarwal2023dexterous} suggests that a blind grasping policy, relying solely on robot proprioception, can robustly grasp unseen objects placed close to the palm. This is because the policy does not overfit to specific object information, leveraging feedback from joint positions and fingertip forces to adapt to various object geometries inherently. Given this insight, we propose training geometry-unaware base policies that only observe proprioception and the 3D positions of objects to infer the object location. Experimental results demonstrate that our geometry-unaware policy, even trained on a single object, generalizes better to a broad range of objects compared to policies with full object perception.

To enhance the diversity of grasping poses across various objects, we introduce a mixture-of-experts (MoE) approach that learns multiple base policies to represent different grasping styles. We use geometric clustering to categorize all objects and train a geometry-unaware policy for each cluster's center. In our multi-task learning framework, we train a residual policy that not only outputs residual actions but also assigns weight to each base policy. The final control action for the robot is determined by a weighted sum of the base policies' actions and the residual action. This method effectively diversifies grasping poses by varying the weights for the base policies, thereby adapting to different object geometries.

ResDex achieves state-of-the-art training performance and generalization capabilities, successfully grasping 3,200 objects in DexGraspNet \citep{wang2023dexgraspnet}. It achieves a success rate of 88.8\% across all training objects and exhibits \textbf{no generalization gap} when applied to unseen objects and categories. Additionally, ResDex demonstrates remarkable training efficiency, mastering such a wide range of tasks in only 12 hours on a single NVIDIA RTX 4090 GPU. In our ablation study, we highlight the critical roles of residual policy learning and geometry-unaware experts in enhancing multi-task learning efficiency and generalization. We also demonstrate the importance of the MoE approach in achieving proper grasping poses.

% \red{summarize our contribution in 3 items, including methodology, techniques and results}

Our main contributions can be summarized as follows:
\begin{itemize}
    \item We introduce ResDex, a novel residual policy learning approach that significantly addresses the problem of efficient multi-task learning and generalization for universal dexterous grasping.
    \item Our technical contributions, including residual multi-task RL, geometry-unaware base policies, and a mixture of experts, demonstrate marked improvements in developing a universal grasping policy.
    \item ResDex achieves state-of-the-art performance on the DexGraspNet dataset, demonstrating its superior training performance and generalization capabilities compared to existing methods.
\end{itemize}

\section{Related Work}
% \red{About 2/3 page in total.}

\textbf{Dexterous Grasping} \citep{multifingered-review, tactile-dex} continues to be a formidable challenge, given the high degrees of freedom in multi-fingered robotic hands and the complex geometries and physical properties of real-world objects.
A fundamental task in dexterous grasping is to generate grasping poses. Recent studies have employed various methods such as contact points \citep{shao2020unigrasp, wu2022learning}, affordance maps \citep{brahmbhatt2019contactgrasp, jiang2021hand}, natural hand annotations \cite{wei2023generalized, hang2024dexfuncgrasp}, and grasping datasets \citep{dexycb, wang2023dexgraspnet} to train models for synthesizing hand grasping poses. 
While generating target grasping poses is crucial, successfully completing a grasp also requires close-loop policies that can manage the entire trajectory.
In learning dexterous grasping policies, both imitation learning \citep{qin2022dexmv, mandikal2022dexvip} and reinforcement learning (RL) \citep{rajeswaran2017learning, wu2024learning} have shown promise. The latter offers scalable advantages across a variety of objects due to its independence from human data collection and the efficiency of simulation environments \citep{makoviychuk2021isaac}.
Recent advancements in research explore universal dexterous grasping using RL for thousands of objects. UniDexGrasp \citep{xu2023unidexgrasp} and UniDexGrasp++ \citep{wan2023unidexgrasp++} introduce curriculum learning and a teacher-student framework to enable training on numerous objects. UniDexFPM \citep{wu2024unidexfpm} extends these approaches to universal functional grasping tasks.
In our study, we propose an improved RL method for universal dexterous grasping that is more efficient and demonstrates superior performance and generalizability.

%UniDexGrasp \citep{xu2023unidexgrasp} uses object curriculum learning and staged training policy in order to get a general policy on objects with different categories, which consumes a lot of training time. 
%UniDexGrasp++ \citep{wan2023unidexgrasp++} uses geometric features and iterative distillation policy and the pipeline is too complicated and cumbersome. 
%\cite{wu2024unidexfpm} adopt a diffusion policy to distill the policies for different categories of objects to get a generalized policy. 
%Different from above existing methods, our work just requires less training process and ensembles fewer policies efficiently.

% Early methods \citep{sahbani2012overview, rosales2012synthesis,liu2021synthesizing} employed analytical methods to achieve a force-closure state.Diffusion models\citep{weng2024dexdiffuser, zhang2024dexgrasp} and are applied in grasp generation. \cite{xu2024dexterous} propose a framework for dexterous grasp generation using Transformer with only one forward pass. In this work, we fully utilize the generated hand poses as a prior to train a general grasping policy. Leveraging a teacher-student distillation \citep{chen2022system, chen2023visual, xu2023unidexgrasp} can obtain a vision-based policy from state-based policy.

% \red{Summarize existing works for scaling up multi-task dex grasping policies. Tell why universal grasping (diverse objects) is difficult? Distinguish our paper from existing approaches.}

\textbf{Residual Policy Learning} provides an effective approach to learn challenging RL tasks when a base policy is available. In robotics, residual policy learning is extensively applied in both manipulation \citep{alakuijala2021residual, davchev2022residual, schoettler2020deep} and navigation tasks \citep{rana2020residual}. 
Typically, the residual policy is constructed upon base policies that employ classical model-based control methods \citep{johannink2019residual, silver2018residual}. 
\citet{garcia2020physics} investigates residual policy learning based on human data.  GraspGF \citep{wu2024learning} explores residual policy learning on a pre-trained score-based generative model \citep{score-matching-denoising}. \citet{zhang2023efficient} and \citet{jiang2024transic} explore using residual policy learning to finetune RL policies. \citet{barekatain2019multipolar} extends residual policy learning to adaptively reweight multiple expert policies. 
In our work, we adopt residual policy learning to tackle the challenges in universal dexterous grasping. Our method, which integrates residual RL with a mixture of geometry-unaware experts, significantly improves multi-task learning to grasp diverse objects.

\textbf{Mixture-of-Experts (MoE)} is initially introduced by \citet{jacobs1991adaptive, jordan1994hierarchical} and typically comprises a set of expert models alongside a gating network \citep{shazeer2017outrageously, fedus2022switch} that learns to weight the output of each expert. Recently, the MoE framework has gained substantial interest in fields such as natural language processing \citep{jiang2024mixtral} and multi-modal learning \citep{mckinzie2024mm1}. MoE has also been applied in RL policies \citep{doya2002multiple, peng2019mcp}, where each expert policy learns a distinct probability distribution that is subsequently integrated. Recent works \citep{cheng2023multi,celik2024acquiring} use MoE to enhance multi-task learning in robotics. 
In our research, we use the MoE framework to improve the diversity of grasping poses in the multi-task learning of dexterous grasping policies. Each expert within our framework is a geometry-unaware policy, trained on an individual object to develop a unique grasping style and achieve broad generalization across a variety of objects.

\section{Preliminaries}

\subsection{Problem Formulation}

We consider tabletop grasping tasks using a 5-fingered ShadowHand to grasp and lift objects initially placed on a table. The hand features 18 DoFs that control a total of 22 joints, including 4 coupled joints. Our goal is to enable grasping any object within a large object set, denoted as $\omega\in\Omega$. For each object, the task is formulated as a Partially Observable Markov Decision Process (POMDP) $M^{\omega} = \langle \mathcal{O}, \mathcal{S}, \mathcal{A}, \mathcal{T}, \mathcal{R}, \mathcal{U} \rangle$, representing the observation space $\mathcal{O}$, the state space $\mathcal{S}$, the action space $\mathcal{A}$, the transition dynamics $\mathcal{T}(s_{t+1}|s_t,a_t)$, the reward function $\mathcal{R}(s_t,a_t)$, and the observation emission function $\mathcal{U}(o_t|s_t)$, respectively. At each timestep $t$, the agent observes $o_t\in \mathcal{O}$ and takes an action $a_t\in \mathcal{A}$, then receives a reward $r_t=\mathcal{R}(s_t,a_t)$. The environment then transitions to the next state $s_{t+1}\sim \mathcal{T}(s_{t+1}|s_t,a_t)$. The agent's objective is to maximize the expected return across all objects $\sum_{\omega\in\Omega} \mathbb{E}\left[ \sum_{t=0}^{T-1} \gamma^t r_t \right]$, where $T$ is the time limit and $\gamma$ is the discount factor.

For task learning in simulation, the observation $o\in\mathcal{O}$ includes: (1) Robot proprioception $\bm{J}\in\mathbb{R}^{123}$, including wrist position and orientation, joint positions of the hand, fingertip states and forces on fingertip sensors; (2) Object pose, including position $\bm{b}^p\in\mathbb{R}^3$ and quaternion $\bm{b}^q\in\mathbb{R}^4$; (3) An object code $\bm{c}^\omega\in\mathbb{R}^{64}$, representing the object geometry via a pre-trained PointNet \citep{pointnet}.  In real-world settings, while precise object pose is unavailable, we opt to use the object point cloud $\bm{p}\in\mathbb{R}^{N\times 3}$,  which contains $N$ points captured by cameras. 
The action $a\in\mathcal{A}$ consists of target joint positions of the hand and the 6D force applied at the wrist.
Our aim is to learn a {vision-based policy} $\pi_\theta^V\left( a_t|\bm{J}_t, \bm{p}_t, a_{t-1} \right)$, parameterized by $\theta$, to maximize the expected return across all objects.

DexGraspNet \citep{wang2023dexgraspnet} provides a dataset that associates each object with grasping proposals. Each grasping proposal is defined as a triplet $\bm{g}=(R, \bm{t}, \bm{q})$, representing the wrist's relative rotation $R\in \mathbb{SO}(3)$ and position $\bm{t}\in\mathbb{R}^3$ to the object and the hand's joint positions $\bm{q}\in\mathbb{R}^{22}$ for a successful grasp. Following \citet{xu2023unidexgrasp}, these data can be integrated into the reward function to facilitate policy learning:
\begin{align}
    & r_t = r_t^{task} + \alpha r_t^{proposal}, \\
    & r_t^{proposal} = -\|\bm{g}-\bm{g}_{t}\|,
\end{align}
where $r_t^{task}$ is a predefined reward for the grasping tasks as detailed in \Cref{appendix:setup}. The reward term $r_t^{proposal}$ penalizes the distance to the grasping proposal, where $\alpha$ is a hyperparameter adjusting its weight and $\bm{g}_t=(R_t, \bm{t}_t, \bm{q}_t)$ represents the current relative pose of the hand to the object.
 
\subsection{The Teacher-Student Framework for Universal Dexterous Grasping}
%\red{Outline: (1) Define the process to first learn state-based policies $\pi^S_{\phi}\left( a_t|\bm{J}_t, \bm{b}^p_t, \bm{b}^q_t, \bm{c}^\omega, a_{t-1} \right)$, then distill into $\pi_\theta^V$. (2) Claim the challenge to learn $\pi^S_{\phi}$ directly on all tasks. (3) Mention existing approaches that learn $\pi^S_{\phi}$ using either complex curriculum or individual training on many clusters.}

Directly optimizing the vision-based policy using RL faces challenges due to gradient interference \citep{gradient-surgery} in multi-task RL and the high dimensionality of point cloud observations. Recent works \citep{xu2023unidexgrasp, wan2023unidexgrasp++, wu2024unidexfpm} have adopted a teacher-student framework in two stages to address these issues. First, a state-based policy $\pi^S_{\phi}\left( a_t|\bm{J}_t, \bm{b}^p_t, \bm{b}^q_t, \bm{c}^\omega, a_{t-1} \right)$ is trained using privileged object information to master all tasks. Then, this policy is distilled into a vision-based policy using DAgger \citep{dagger}, an online imitation learning method. 

To address the multi-task optimization challenge in learning the state-based policy, UniDexGrasp \citep{xu2023unidexgrasp} proposed a curriculum learning approach. The RL training starts with a single object and, after a certain number of iterations, gradually includes more objects. This process continues until all objects are included and the policy achieves a high success rate. 
UniDexGrasp++ \citep{wan2023unidexgrasp++} introduced an improved method based on generalist-specialist learning \citep{gsl}. The entire object set is divided into groups through geometry-aware clustering. Numerous specialist state-based policies are then trained and subsequently distilled into a generalist state-based policy, with iterative training implemented through a curriculum.
These methods require meticulous curriculum design and are time-consuming, as various policies are trained across different sets of objects. Additionally, their learned vision-based policies exhibit a significant decrease of about $7\%$ in success rates when tested on unseen objects, indicating limited generalization capabilities.

\section{Method}

\begin{figure}[!t]
  \centering
  \includegraphics[width=.95\linewidth, trim={0cm, 10.5cm, 4cm, 0cm}, clip]{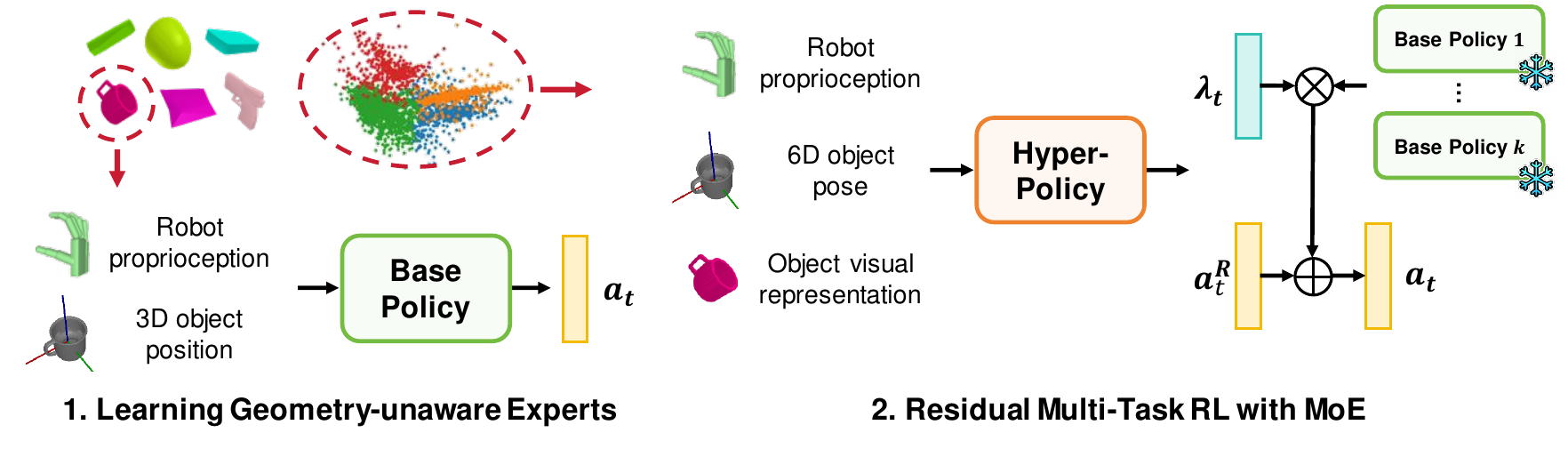}
  \vspace{0mm}
  \caption{We propose ResDex, an efficient learning framework for dexterous grasping across thousands of objects. The learning process consists of two stages: (1) For each representative object from the cluster center, we train a geometry-unaware base policy, which provides weak generalization across a broad range of objects. (2) To develop a universal policy applicable to all objects, we use residual multi-task reinforcement learning (RL) to train a hyper-policy, incorporating the base policies within a mixture-of-experts (MoE) framework. ResDex demonstrates efficient training and robust generalization to unseen objects. }
  \label{fig:framework}
\end{figure}

We propose ResDex, a framework that leverages residual policy learning combined with a mixture of experts to provide an efficient approach for universal dexterous grasping, significantly enhancing generalization capabilities. Figure \ref{fig:framework} illustrates an overview of our framework.

\subsection{Learning Geometry-Unaware Policies}
\label{sec:blindgrasp}

To enable efficient multi-task RL using residual policy learning, it is essential to build a base policy that can effectively explore all the involved tasks. Training a base policy directly on a single type of object often results in overfitting, which significantly decreases its generalizability to other objects. 
Conversely, training a policy on all objects using RL presents unique challenges, as different tasks can lead to gradient interference in the learning processes, making the training highly inefficient.

We propose to build base policies that, while trained on a limited number of objects, can generalize effectively to a wider range of objects. 
Empirical insights from \citet{agarwal2023dexterous} suggest that a blind grasping policy, trained solely on robot proprioception without specific object information, can better generalize to unseen objects. 
We hypothesize that limiting observations helps the policy avoid overfitting to specific object features. When a policy does not have complete information about object poses and geometric features, it tends to learn more generalizable grasping strategies and rely on the proprioceptive feedback to adjust actions.
Although we cannot use a fully blind policy in our setting -- as the agent must know the object's location to approach it -- we integrate this insight by proposing a \textbf{geometry-unaware base policy}, $\pi^B_{\psi}\left( a_t|\bm{J}_t, \bm{b}^p_t, a_{t-1} \right)$, which uses only robot proprioception $\bm{J}$ and the 3D position of the object $\bm{b}^p$.

The grasping proposal reward $r^{proposal}$ inherently leaks the object's geometric information, as the target relative wrist pose specifies ``where to grasp on the object''. To mitigate this unwanted information leakage and enhance generalization, we replace this term with a pose reward:
\begin{equation}
    r^{pose}_t = -\|\bm{q}-\bm{q}_{t}\|,
\end{equation}
where $\bm{q}_{t}$ represents the current hand joint positions. This reward encourages the hand to reach the target joint positions, focusing on the hand pose rather than the specific region to grasp on the object.

Experimental results (Section \ref{sec:ablation}) show that our geometry-unaware policy, trained on a single object, demonstrates remarkable generalizability to unseen objects and significantly outperforms policies that incorporate full observations or those trained using the full grasping proposal reward.

\subsection{Residual Multi-Task Reinforcement Learning}\label{sec:residual}
%\red{present the idea of using generalizable base policy + residual RL for efficient multi-task learning. showcase the residual learning method given one base policy}

%Directly learning a general grasping policy on the training set is theoretically challenging due to multi-task gradient interference \citep{gradient-surgery}, and it is not feasible in practice. 
%Although it is possible to learn a general grasping policy using a teacher-student framework, this approach is time-consuming and complex.

%To overcome these challenges and directly learn a multi-task dexterous grasping policy across thousands of objects, we combine a generalizable base policy with residual reinforcement learning. 
%This combination facilitates efficient multi-task learning for universal dexterous grasping while enhancing overall generalizability.

While the base policy trained on a single object offers some degree of generalizability across various objects, it typically achieves a low overall success rate. To address this, we introduce residual policy learning to develop a policy that masters all objects.

The state-based residual policy, denoted as $\pi^R_{\phi}\left( a_t|\bm{J}_t, \bm{b}^p_t,\bm{b}^q_t, \bm{c}^\omega, a_{t-1} \right)$, is parameterized by $\phi$. It utilizes all available state-based observations to better maximize performance in solving POMDPs. 

Given the pre-trained base policy $\pi^B_{\psi}$, at each timestep, the base policy uses the required observations from the complete observations to compute a base action $a_{t}^B = \argmax_{a_t} \pi^B_{\psi}\left( a_t|\bm{J}_t, \bm{b}^p_t, a_{t-1} \right)$. Simultaneously, the residual policy samples a residual action $a_t^R\sim\pi^R_{\phi}\left( a_t|\bm{J}_t, \bm{b}^p_t,\bm{b}^q_t, \bm{c}^\omega, a_{t-1} \right)$, and these actions are combined element-wise to form the final action $a_t=a_t^B+a_t^R$.

The generalizability of the base policy reduces the need for extensive exploration by the residual policy across diverse object geometries, making it practical to train under multi-task settings. For objects already successfully grasped by the base policy, the residual policy refines the grasping process, enhancing the success rate. For objects not successfully grasped by the base policy, the residual policy can efficiently explore in the residual action space, benefiting from the significant exploration bias provided by the base policy. 
We train this residual policy across the entire object set using RL, aiming to maximize the average return across all objects. Our experiments demonstrate that a single base policy, when aided by the residual policy, can achieve high success rates across thousands of objects.

% \begin{equation}
%     \begin{aligned}
%         &a = a_{base} + a_{resl} \\
% &a_{base}\sim\pi^S_{\theta}\left( a_t|\bm{J}_t, \bm{b}^p_t, a_{t-1} \right) \\
% &a_{res}\sim \pi^S_{\phi}\left( a_t|\bm{J}_t, \bm{b}^p_t,\bm{b}^q_t, a_{t-1}, \bm{c}^\omega\right)
%     \end{aligned}
% \end{equation}

%Despite significant variations in object geometry, we assume that feasible grasping poses tend to be closely clustered. 

\subsection{Incorporating a Mixture of Experts}
\label{sec:moe}
%\red{Give the motivation of combining several poses for universal grasping. Present the MoE part.}

Utilizing diverse poses to grasp different objects is not only a crucial feature for dexterous hands but also essential for post-grasping manipulations in real-world tasks.
While residual policy learning based on a single base policy can achieve commendable success rates, it often struggles to perform various grasping poses for different objects. This limitation arises because the base policy typically provides only a single grasping pose for its training object, thus posing a significant challenge for the residual policy to explore diverse grasping poses for certain objects.

To enhance the diversity of grasping poses, we propose a mixture-of-experts (MoE) approach. In this setup, several base policies are trained, each capable of executing distinct grasping styles, and their actions can be combined to generate a variety of novel grasping poses.
To acquire base policies that exhibit diverse behaviors and grasping poses, we use geometry-aware clustering \citep{wan2023unidexgrasp++} to divide the object set into $k$ clusters based on object shape representations.  Objects at the cluster centers are used to train the base policies $\{\pi^B_{\psi_i}\}_{i=1}^k$, leveraging their distinct and representative shapes to foster diversified grasping styles.

In multi-task learning, to integrate the base policies while learning residual actions, we replace the residual policy with a \textbf{hyper-policy}, denoted as $\pi^H_{\phi}\left(a^R_t, \bm{\lambda}_t|\bm{J}_t, \bm{b}^p_t,\bm{b}^q_t, \bm{c}^\omega, a_{t-1} \right)$. This hyper-policy predicts the residual action $a_t^R\in\mathcal{A}$ along with a weight $\bm{\lambda}_t\in \mathbb{R}^k$ for the MoE.
At each timestep, all base policies predict base actions $\{a^B_{t,i}\}_{i=1}^k$ using partial observations, and the hyper-policy samples the weights and the residual action. The final action is computed as follows:
\begin{equation}
    a_t = a^R_t + \frac{1}{\|\bm{\lambda}_t\|} \sum_{i=1}^k {\lambda_{t,i}a^B_{t,i}},
\end{equation}
using a normalized weighted sum of base actions in addition to the residual action. This hyper-policy aims at efficiently learning diverse, natural grasping poses by adjusting the MoE weights and enhancing multi-task performance through residual learning.

%While we highlighted the condensed properties of feasible grasping poses in section \ref{sec:residual}, the diversity of these poses remains limited when relying solely on a single base policy. Our experiments demonstrate that MoE significantly improves the policy's learning ability when guided by a goal-conditioned reward function. 
%Furthermore, it plays a crucial role in enabling our framework to learn more natural and proper grasping poses.

\subsection{Method Summary}
%\red{summarize the whole pipeline. Because we give the teacher-student framework in preliminaries, here we can quickly go through the distillation process.}

Here, we outline the complete pipeline to train ResDex, which encompasses three training phases:

\textbf{Training Base Policies: } Using the entire training set of objects, we apply K-Means clustering \citep{kmeans} on the PointNet \cite{pointnet} features of objects to generate  $k$ clusters. From each cluster, we select the object closest to the center and train a geometry-unaware base policy for each object using RL, as detailed in Section \ref{sec:blindgrasp}.

\textbf{Training the Hyper-Policy: } We train the hyper-policy across parallel environments that span all objects in the training set, as described in Section \ref{sec:moe}.
During the training process, the hyper-policy is continually updated while the base policies remain fixed. 
To cultivate diverse and effective grasping poses while maximizing success rates, we employ a two-stage reward function:
\begin{itemize}
    \item \textbf{First stage: } We use the reward function that includes the grasping proposal reward: $r = r^{task} + r^{proposal}$. This reward function guides the policy to follow the reference grasping poses provided by the dataset, resulting in more natural and human-like grasps.
    \item \textbf{Second stage: } We remove the grasping proposal term in the reward function and eliminate terms that encourage approaching the object within $r^{task}$, focusing solely on terms related to object lifting and task completion. This adjustment further enhances the policy's performance by removing constraints imposed by these reward terms. Further details on the reward functions are provided in Appendix \ref{appendix:setup}.
\end{itemize}

\textbf{Vision-based Distillation: } To learn a vision-based policy $\pi_\theta^V$ that operates without privileged object information, we adopt the teacher-student framework. The state-based hyper-policy serves as the teacher, and the vision-based policy to learn acts as the student. We use DAgger \citep{dagger} to train, which involves iteratively collecting trajectories with the student policy and supervising it using the teacher policy.

\section{Experiments}
\subsection{Experiment Settings}
We evaluate the effectiveness of our method on DexGraspNet \citep{wang2023dexgraspnet}, a large-scale robotic dexterous grasping dataset for thousands of everyday objects. 
The dataset is split into one training set and two test sets, including one that contains unseen objects in the seen categories and the other that contains unseen objects in unseen categories. 
The training set includes 3,200 object instances, while the test sets contain a total of 241 object instances.

% We use MLP with 4 hidden layers (1024, 1024, 512, 512) for base policies, hyper policies and vision-based policies. We use PPO to train both the base policies and hyper policies. 
%For the state-based setting, the observation includes robot proprioception $\bm{J}$, object 6D pose $\bm{b}^p$ and $\bm{b}^q$, and preprocessed geometric features $\bm{c}^\omega$.
%For the vision-based setting, the observation includes robot proprioception $\bm{J}$ and object point clouds $\bm{p}^{N\times3}$. 
To train RL policies, we set up parallel simulation environments using IsaacGym \citep{makoviychuk2021isaac}. For vision-based distillation, we sample 512 points on each object's mesh to provide point cloud observations. %points are then processed into geometric features using a PointNet, which is trained concurrently with the student policy during the distillation process.
We compare our ResDex with state-of-the-art methods including UniDexGrasp \citep{xu2023unidexgrasp} and UniDexGrasp++ \citep{wan2023unidexgrasp++}.

\subsection{Main Results}\label{sec:mainresult}
Table \ref{tab:first_training_result} shows that our method outperforms UniDexGrasp++ by 2.7\%, 5.4\%, and 7.8\% respectively on the training set and two test sets when $k = 4$ after the first training stage. Our method consistently outperforms the baselines using $k$ larger than $2$.
Additionally, we notice that increasing $k$ leads to a slight performance gain. %there is a gap in success rates between small and large values of $k$.
This indicates that using a mixture of base policies enables the hyper-policy to better align with the guidance provided by the grasping proposal reward.

\begin{table}[t]
\caption{\textbf{Success rates of state-based policies after the first training stage.} We evaluate our method on three different random seeds. $k$ denotes the number of geometry-unaware base policies used to train the hyper-policy.}
\label{tab:first_training_result}
\begin{center}
    \begin{tabular}{c|c|cc}
\toprule
\multirow{3}{*}{\textbf{Method}} & \multirow{3}{*}{\textbf{Train(\%)}} & \multicolumn{2}{c}{\textbf{Test(\%)}} \\
                                &                                     & \textbf{Uns. Obj.} & \textbf{Uns. Cat.} \\
                                &                                     & \textbf{Seen Cat.} &                    \\
\midrule
UniDexGrasp                     & 79.4                                & 74.3                         & 70.8               \\
UniDexGrasp++                   & 87.9                      & 84.3                & 83.1      \\
\midrule

ResDex ($k=1$)              &  83.2$\pm$ 1.5                             & 82.8$\pm$1.0                         & 85.1$\pm$0.9              \\
ResDex ($k=2$)               & 82.8$\pm$ 3.9                             & 82.6$\pm$3.2                          & 85.0$\pm$3.3              \\
ResDex ($k=3$)               & 88.1$\pm$ 1.2                              & 88.2$\pm$0.4                          & 89.3$\pm$1.0              \\
ResDex ($k=4$)               &\textbf{90.6}$\pm$\textbf{0.6}                             & \textbf{89.7}$\pm$\textbf{0.8}                          & \textbf{90.9}$\pm$\textbf{0.1}              \\
ResDex ($k=5$)               & 87.6$\pm$ 0.5                              & 87.3$\pm$0.8                          & 88.1$\pm$0.2              \\
ResDex ($k=6$)               & 88.7$\pm$ 0.6                              & 87.8$\pm$0.5                          & 88.8$\pm$1.1              \\
\bottomrule
\end{tabular}
\end{center}
\end{table}

\begin{table}[t]
\caption{\textbf{Success rates of state-based policies after the second training stage. } We evaluate our method on three different random seeds. $k$ denotes the number of geometry-unaware base policies used to train the hyper-policy.}
\begin{center}
    \begin{tabular}{c|c|cc}
\toprule
\multirow{3}{*}{\textbf{Method}} & \multirow{3}{*}{\textbf{Train(\%)}} & \multicolumn{2}{c}{\textbf{Test(\%)}} \\
                                &                                     & \textbf{Uns. Obj.} & \textbf{Uns. Cat.} \\
                                &                                     & \textbf{Seen Cat.} &                    \\
\midrule

UniDexGrasp                     & 79.4                                & 74.3                         & 70.8               \\
UniDexGrasp++                   & 87.9                      & 84.3                & 83.1      \\
\midrule
ResDex ($k=1$)              &  94.3$\pm$1.6                               & 93.8$\pm$1.8                        &94.5$\pm$1.3               \\
ResDex ($k=2$)               & 94.5$\pm$0.9                              & 94.3$\pm$1.1                          & 95.2$\pm$1.0              \\
ResDex ($k=3$)               & 94.1$\pm$0.9                              & 93.9$\pm$0.9                          & 94.4$\pm$1.2              \\
ResDex ($k=4$)               & \textbf{94.6}$\pm$\textbf{1.6}                              & \textbf{94.4}$\pm$\textbf{1.7}                          & \textbf{95.4}$\pm$\textbf{1.0}              \\
ResDex ($k=5$)               & 94.2$\pm$0.5                              & 93.7$\pm$0.9                          & 94.2$\pm$0.6              \\
ResDex ($k=6$)               & 93.9$\pm$1.3                              & 93.6$\pm$1.6                          & 94.5$\pm$1.1              \\
\bottomrule
\end{tabular}
\end{center}
\label{tab:second_training_result}
\end{table}

Table~\ref{tab:second_training_result} demonstrates that the second training stage significantly boosts the success rates of all policies, regardless of the value of $k$. 
Specifically, our best-performing policy ($k=4$) outperforms UniDexGrasp++ by 6.7\%, 10.1\%, and 12.3\% on the training and test sets respectively.
Unlike previous methods, our approach shows no generalization gap, achieving consistent success rates on both the training and test sets.
This consistency indicates that our method can provide a grasping policy that is more robust and generalizable.

During distillation, we use the hyper-policy trained with four base policies as the teacher to learn a vision-based policy. 
Performance of vision-based policies are presented in Table~\ref{tab:vision-based result}. Our vision-based policy outperforms UniDexGrasp++ by 3.4\%, 8.9\% and 10.5\% in success rates on the three object sets respectively, demonstrating strong generalization capabilities to unseen objects.

\begin{table}[!ht]
    \centering
    \caption{\textbf{Success rates of vision-based policies.} %We distill the hyper-policy with 4 pretrianed base policies into a vision-based policy.
    }
    \begin{tabular}{c|c|cc}
    \toprule
\multirow{3}{*}{\textbf{Methods}} & \multirow{3}{*}{\textbf{Train(\%)}} & \multicolumn{2}{c}{\textbf{Test(\%)}} \\
                                &                                     & \textbf{Uns. Obj.} & \textbf{Uns. Cat.} \\
                                &                                     & \textbf{Seen Cat.} &                    \\
\midrule
UniDexGrasp     & 73.7 &68.6 & 65.1 \\
UniDexGrasp++   & 85.4 &79.6 & 76.7 \\
\midrule
ResDex ($k=4$)       & \textbf{88.8} &\textbf{88.5} & \textbf{87.2} \\
\bottomrule
    \end{tabular}
    \label{tab:vision-based result}
\end{table}

\subsection{Ablation study} \label{sec:ablation}

\begin{figure}[t]
    \centering
    \begin{minipage}[b]{0.3\textwidth}
        \centering
        \setlength{\abovecaptionskip}{0mm}
        \includegraphics[width=\linewidth]{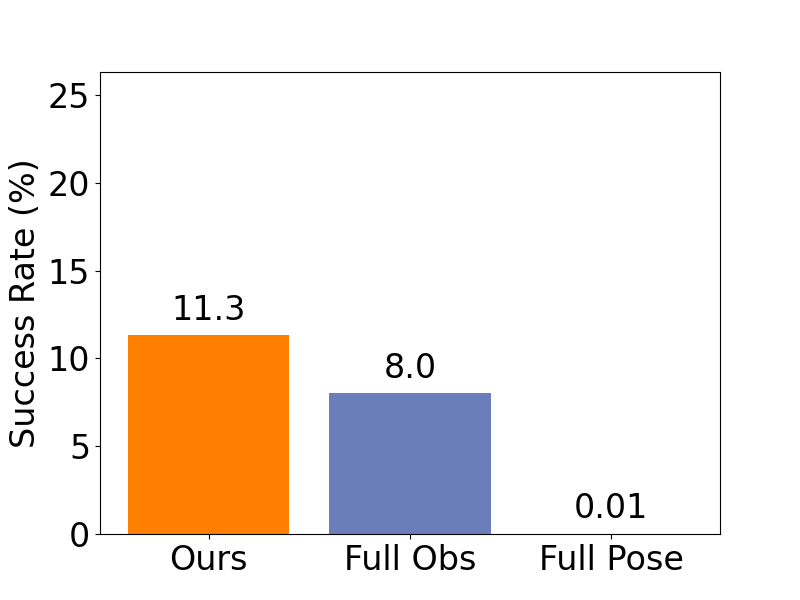}
        \caption*{Cell Phone}
        \label{fig:image1}
    \end{minipage}
    %\hspace{0.3cm}
    \begin{minipage}[b]{0.3\textwidth}
        \centering
        \setlength{\abovecaptionskip}{0mm}
        \includegraphics[width=\linewidth]{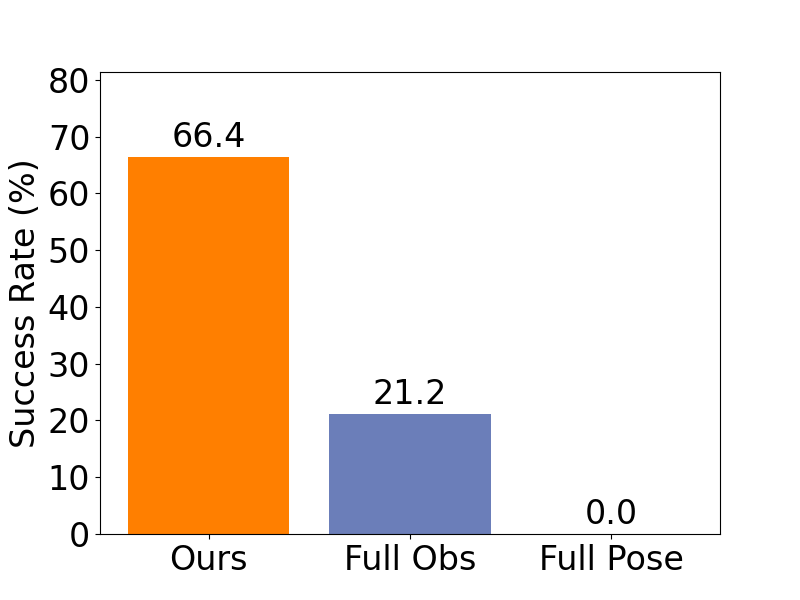}
        \caption*{Toy Figure}
        \label{fig:image2}
    \end{minipage}
    %\hspace{0.3cm}
    \begin{minipage}[b]{0.3\textwidth}
        \centering
        \setlength{\abovecaptionskip}{0mm}
        \includegraphics[width=\linewidth]{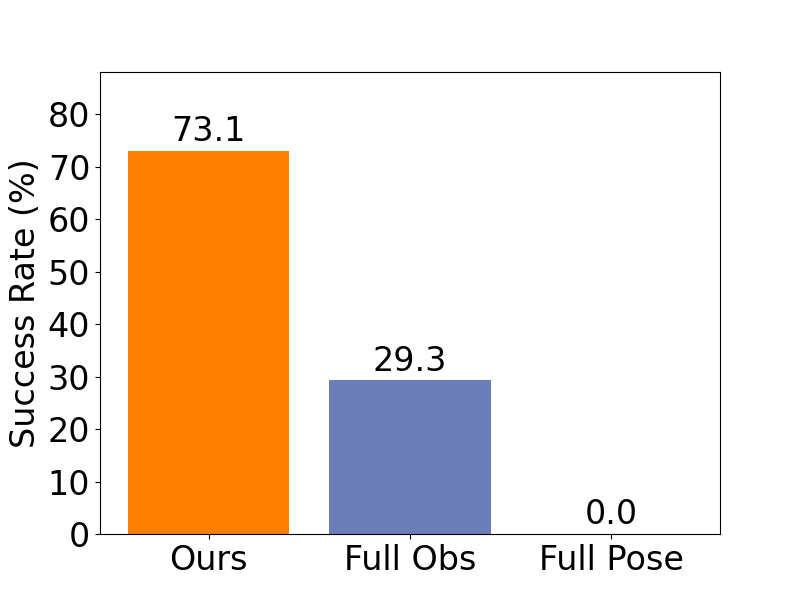}
        \caption*{Bottle}
        \label{fig:image3}
    \end{minipage}
    
    \vspace{0em}  % Space between rows

    \begin{minipage}[b]{0.3\textwidth}
        \centering
        \setlength{\abovecaptionskip}{0mm}
        \includegraphics[width=\linewidth]{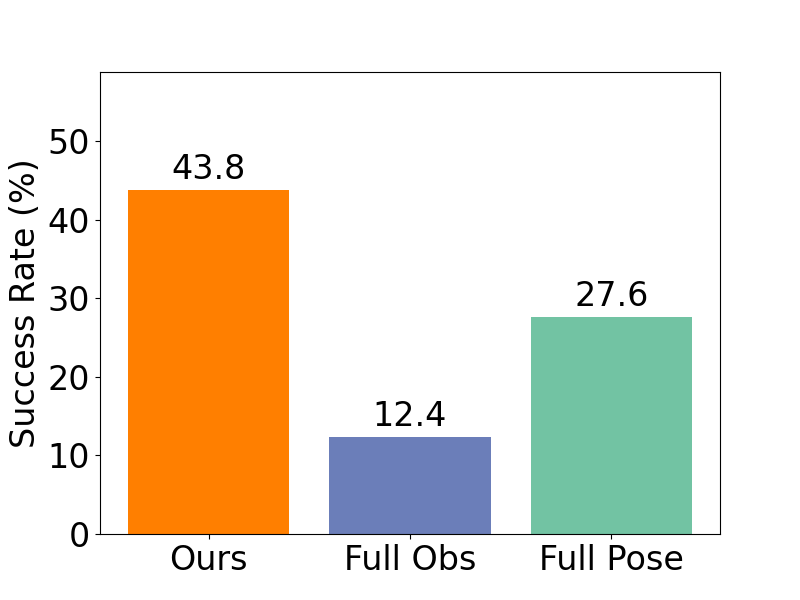}
        \caption*{Video Game Console}
        \label{fig:image4}
    \end{minipage}
    %\hspace{0.3cm}
    \begin{minipage}[b]{0.3\textwidth}
        \centering
        \setlength{\abovecaptionskip}{0mm}
        \includegraphics[width=\linewidth]{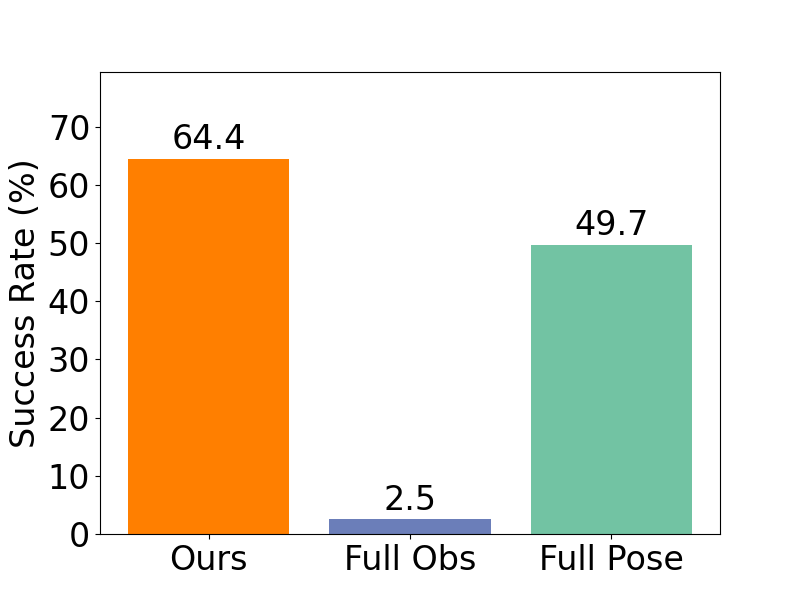}
        \caption*{Toilet Paper}
        \label{fig:image5}
    \end{minipage}
    %\hspace{0.3cm}
    \begin{minipage}[b]{0.3\textwidth}
        \centering
        \setlength{\abovecaptionskip}{0mm}
        \includegraphics[width=\linewidth]{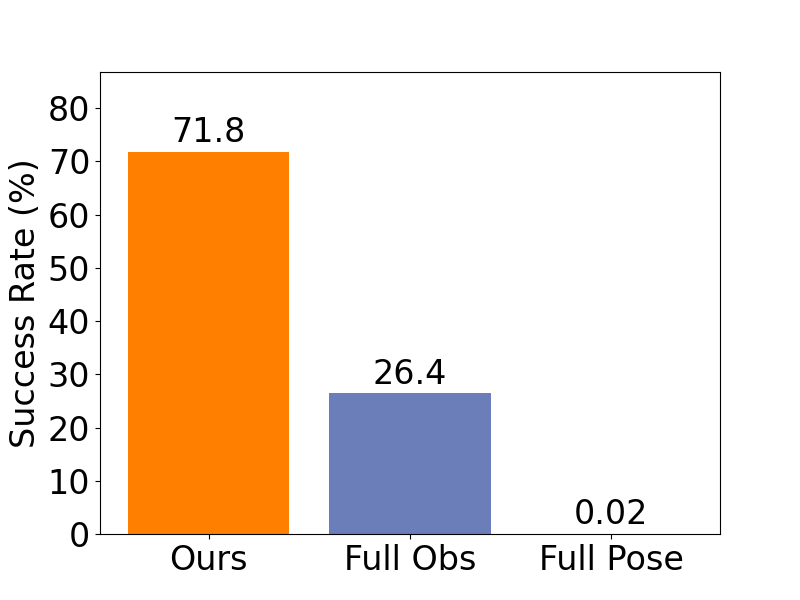}
        \caption*{Mug}
        \label{fig:image6}
    \end{minipage}
    
    \caption{\textbf{Generalization performance to all objects using policies with different observations and reward, each trained on a single object.} Ours: Geometry-unaware policy. Full Obs: Policy trained with the complete state-based observations. Full Pose: Policy trained using the reward function that includes the full grasping proposal reward. %``Full pose'' demonstrates  0\% success rates for the cell phone, toy figure, bottle and mug due to limited observation. 
    }
    \label{fig:policy comparare}

\end{figure}

\textbf{Geometry-Unaware Experts.} \ We compare generalizability between geometry-unaware policies and policies trained with full state-based observations. We train 3 types of policies on 6 objects, including cell phone, toy figure, bottle, video game console, toilet paper and mug, and we evaluate their performance on the training set. The results are shown in Figure \ref{fig:policy comparare}. Our geometry-unaware policies achieve higher success rates compared to other policies, achieving over 70\% success rates when trained on some objects, which demonstrates remarkable generalizability.
Policies with the full observations or the full grasping proposal reward demonstrate poor generalization when trained on some specific objects.

% old

% old 
\begin{figure}[t]
    % First image with caption but not counted
    \centering
    \begin{minipage}{0.19\textwidth}
        \centering
        \includegraphics[width=\textwidth]{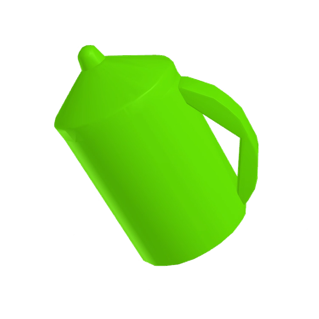} % Change to your image path
        % \caption*{Kettle}
    \end{minipage}
    %\hfill
    % Row of images without captions
    \begin{minipage}{0.19\textwidth}
        \setlength{\abovecaptionskip}{1mm}
        \centering
        \caption*{$k=1$}
        \includegraphics[width=\textwidth]{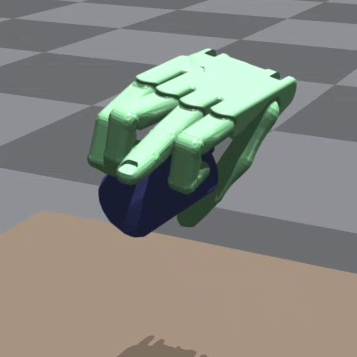} % Change to your image path
    \end{minipage}
    %\hfill
    \begin{minipage}{0.19\textwidth}
        \centering
        \setlength{\abovecaptionskip}{1mm}
        \caption*{$k=2$}
        \includegraphics[width=\textwidth]{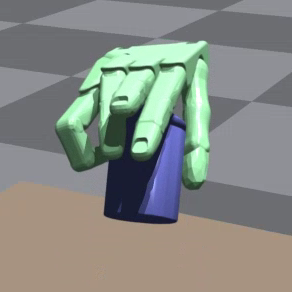} % Change to your image path
    \end{minipage}
    %\hfill
    \begin{minipage}{0.19\textwidth}
        \centering
        \setlength{\abovecaptionskip}{1mm}
        \caption*{$k=3$}
        \includegraphics[width=\textwidth]{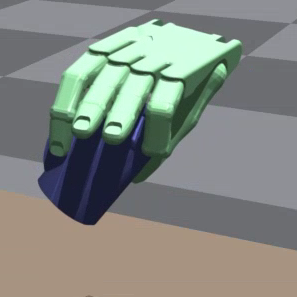} % Change to your image path
    \end{minipage}
    %\hfill
    \begin{minipage}{0.19\textwidth}
        \centering
        \setlength{\abovecaptionskip}{1mm}
        \caption*{$k=4$}
        \includegraphics[width=\textwidth]{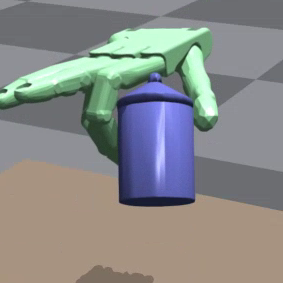} % Change to your image path
    \end{minipage}
    
    \vspace{0em} % Space between the rows

    % Second row of images
    \begin{minipage}{0.19\textwidth}
        \centering
        \includegraphics[width=\textwidth]{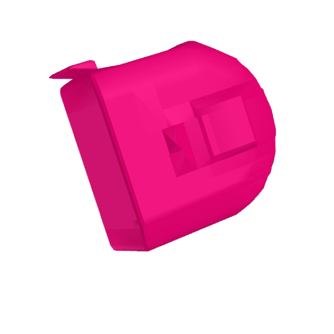} % Change to your image path
        % \caption*{Tape Measure}
    \end{minipage}
    %\hfill
    \begin{minipage}{0.19\textwidth}
        \centering
        \includegraphics[width=\textwidth]{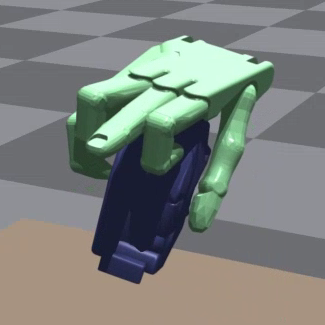} % Change to your image path
    \end{minipage}
    %\hfill
    \begin{minipage}{0.19\textwidth}
        \centering
        \includegraphics[width=\textwidth]{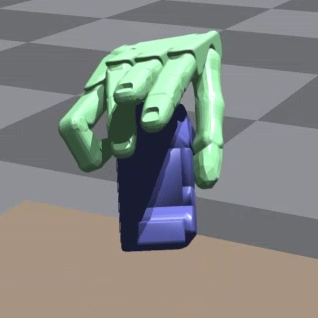} % Change to your image path
    \end{minipage}
    %\hfill
    \begin{minipage}{0.19\textwidth}
        \centering
        \includegraphics[width=\textwidth]{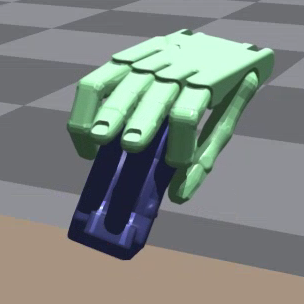} % Change to your image path
    \end{minipage}
    %\hfill
    \begin{minipage}{0.19\textwidth}
        \centering
        \includegraphics[width=\textwidth]{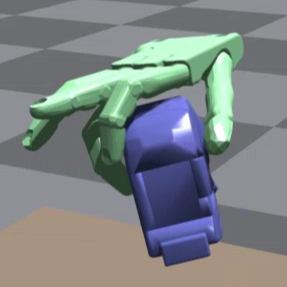} % Change to your image path
    \end{minipage}
 
    \vspace{0em} % Space between the rows

    % Third row of images
    \begin{minipage}{0.19\textwidth}
        \centering
        \includegraphics[width=\textwidth]{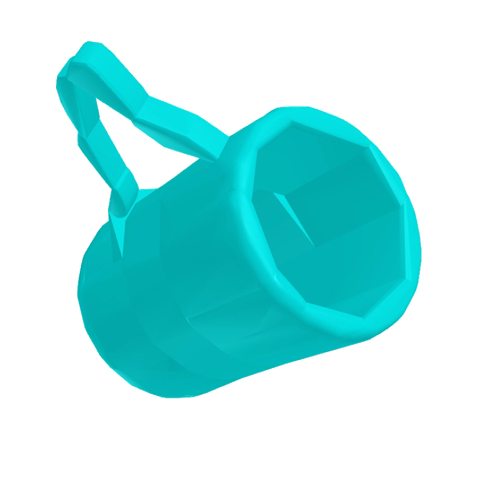} % Change to your image path
        % \caption*{Mug}
    \end{minipage}
    %\hfill
    \begin{minipage}{0.19\textwidth}
        \centering
        \includegraphics[width=\textwidth]{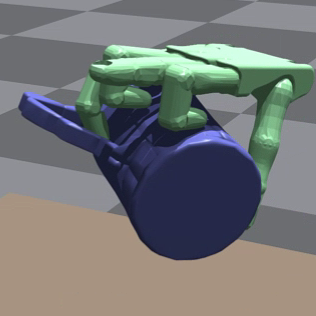} % Change to your image path
    \end{minipage}
    %\hfill
    \begin{minipage}{0.19\textwidth}
        \centering
        \includegraphics[width=\textwidth]{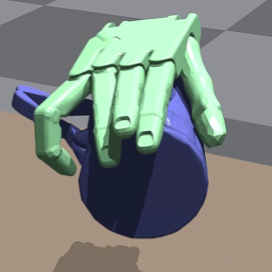} % Change to your image path
    \end{minipage}
    %\hfill
    \begin{minipage}{0.19\textwidth}
        \centering
        \includegraphics[width=\textwidth]{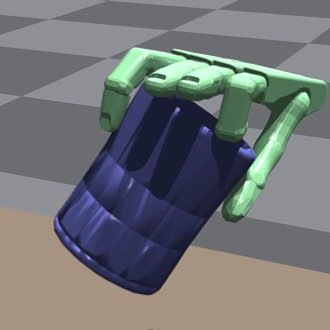} % Change to your image path
    \end{minipage}
    %\hfill
    \begin{minipage}{0.19\textwidth}
        \centering
        \includegraphics[width=\textwidth]{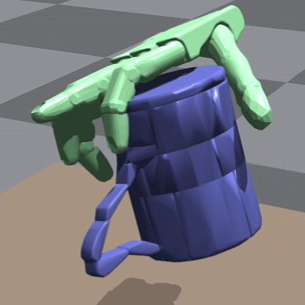} % Change to your image path
    \end{minipage}

    \vspace{0em} % Space between the rows

    % fourth row of images
    \begin{minipage}{0.19\textwidth}
        \centering
        \includegraphics[width=\textwidth]{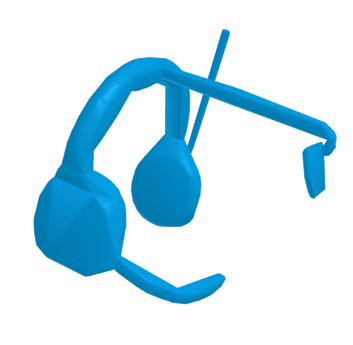} % Change to your image path
    \end{minipage}
    %\hfill
    \begin{minipage}{0.19\textwidth}
        \centering
        \includegraphics[width=\textwidth]{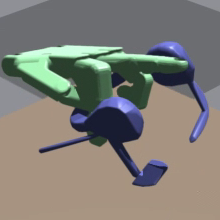} % Change to your image path
    \end{minipage}
    %\hfill
    \begin{minipage}{0.19\textwidth}
        \centering
        \includegraphics[width=\textwidth]{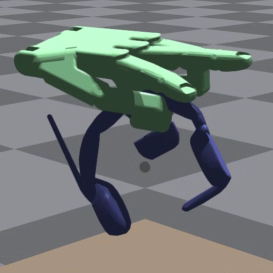} % Change to your image path
    \end{minipage}
    %\hfill
    \begin{minipage}{0.19\textwidth}
        \centering
        \includegraphics[width=\textwidth]{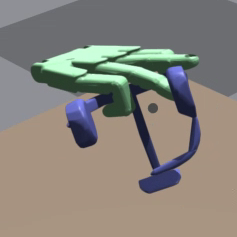} % Change to your image path
    \end{minipage}
    %\hfill
    \begin{minipage}{0.19\textwidth}
        \centering
        \includegraphics[width=\textwidth]{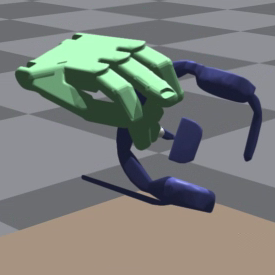} % Change to your image path
    \end{minipage}
    
    \caption{\textbf{Grasping poses achieved by hyper-policies trained with various numbers of base policies.} Each row displays grasping poses for a kettle, tape measure, mug, and headphone, respectively. Columns show hyper-policies trained with 1, 2, 3, and 4 base policies, arranged from left to right. }
    \label{fig:grasping pose}
\end{figure}

\begin{table}[!ht]
\caption{\textbf{Ablation study on residual reinforcement learning.} We assess success rates of policies on the training set. \textbf{Method} indicates the number of base policies used. \textbf{MoE} shows the results for hyper-policies without residual actions, while \textbf{MoE+Res} shows the results for policies that output both normalized weights for MoE and residual actions. }
\begin{center}
    \begin{tabular}{ccccccc}
\toprule
\textbf{Method} &  $k=1$ &$k=2$ &$k=3$ & $k=4$ & $k=5$ &$k=6$ \\
\midrule
\textbf{MoE} & 61.4 &71.1 &79.4 &80.3 &72.1 &81.6 \\
%\midrule
\textbf{MoE+Res}&83.2 &82.8 &88.1 &90.6 &87.6& 88.7\\
\bottomrule
\end{tabular}
\end{center}

\label{tab:residual_ablation}
\end{table}

\textbf{Residual Reinforcement Learning.} \ To demonstrate the multi-task learning ability provided by residual reinforcement learning, we implement an ablation method that combines base policies using a hyper-policy which only outputs the weights without residual actions. We evaluate the performance on the training set. The results, as shown in Table \ref{tab:residual_ablation}, demonstrate that for different number of base policies, the method with residual learning can notably boost the performance.

\textbf{Mixture-of-Experts.} \ We further demonstrate that a mixture of base policies can generate better grasping poses. 
We assess the quality of the grasping poses executed by our policies  by  computing $D = -\sum_{t=1}^T r_t^{proposal}$. The term $r_t^{proposal}$ is a negative reward that punishes the difference between the current grasping pose and the grasping proposal $(R, \bm{t}, \bm{q})$. Therefore, the higher the value of $D$, the less natural the grasping poses executed by the policy.
The results, as shown in Table~\ref{tab:naturalness}, reveal that for the hyper-policies trained over two stages, although their success rates are very close, those with more base policies generally display better grasping poses.

\begin{table}[!ht]
\caption{\textbf{Quality of grasping poses achieved by different policies.} We evaluate the $D$ values of ResDex policies with various $k$ on the test set of unseen objects in unseen categories. The lower $D$ means the better grasping poses achieved.
%$k$ means we pretrain $k$ geometry-unaware policies for the hyper policy.
}
\begin{center}
    \begin{tabular}{ccccccc}
\toprule 
\textbf{Methods} & $k=1$ & $k=2$ & $k=3$ & $k=4$ & $k=5$ & $k=6$  \\
\midrule
$\boldsymbol{D}\downarrow$  & 223.6 & 174.5 & 194.3 & 176.3 & 204.6 & 176.1 \\

\bottomrule
\end{tabular}   
\end{center}
\label{tab:naturalness}
\end{table}

Moreover, Figure \ref{fig:grasping pose} illustrates the various grasping poses executed by our policies with different numbers of base policies on randomly selected objects (kettle, tape measure, mug, and headphones). 
We observe that ResDex trained with more base policies tend to learn grasping strategies that are more appropriate and natural.

\section{Conclusion and Limitations}

We propose ResDex for universal dexterous grasping, a framework that effectively addresses the challenges of training efficiency and generalization that are prevalent in existing methods. Our technical contributions include a residual policy learning framework designed for efficient multi-task reinforcement learning in dexterous grasping, a method to train geometry-unaware base policies that enhances generalization and facilitates exploration across multiple tasks, and an MoE framework that enriches the diversity of the learned grasping poses. We demonstrate the superior performance of ResDex compared to existing methods on the large-scale object dataset DexGraspNet, notably achieving a zero generalization gap to unseen objects. The framework also showcases promising simplicity and training efficiency, marking a significant step towards scaling up dexterous learning.

The limitations of our work include: (1) Although we incorporate a grasping proposal reward to refine grasping poses, we have not yet considered the task as functional grasping. Future work could extend our approach to functional grasping tasks to further enhance general robotic manipulation in real-world settings. (2) We have not deployed the vision-based policy on hardware. Future efforts should focus on this aspect and overcome the sim-to-real gap.

\bibliography{iclr2025_conference}

\begin{thebibliography}{49}
\providecommand{\natexlab}[1]{#1}
\providecommand{\url}[1]{\texttt{#1}}
\expandafter\ifx\csname urlstyle\endcsname\relax
  \providecommand{\doi}[1]{doi: #1}\else
  \providecommand{\doi}{doi: \begingroup \urlstyle{rm}\Url}\fi

\bibitem[Agarwal et~al.(2023)Agarwal, Uppal, Shaw, and Pathak]{agarwal2023dexterous}
Ananye Agarwal, Shagun Uppal, Kenneth Shaw, and Deepak Pathak.
\newblock Dexterous functional grasping.
\newblock In \emph{7th Annual Conference on Robot Learning}, 2023.

\bibitem[Alakuijala et~al.(2021)Alakuijala, Dulac-Arnold, Mairal, Ponce, and Schmid]{alakuijala2021residual}
Minttu Alakuijala, Gabriel Dulac-Arnold, Julien Mairal, Jean Ponce, and Cordelia Schmid.
\newblock Residual reinforcement learning from demonstrations.
\newblock \emph{arXiv preprint arXiv:2106.08050}, 2021.

\bibitem[Barekatain et~al.(2019)Barekatain, Yonetani, and Hamaya]{barekatain2019multipolar}
Mohammadamin Barekatain, Ryo Yonetani, and Masashi Hamaya.
\newblock Multipolar: Multi-source policy aggregation for transfer reinforcement learning between diverse environmental dynamics.
\newblock \emph{arXiv preprint arXiv:1909.13111}, 2019.

\bibitem[Brahmbhatt et~al.(2019)Brahmbhatt, Handa, Hays, and Fox]{brahmbhatt2019contactgrasp}
Samarth Brahmbhatt, Ankur Handa, James Hays, and Dieter Fox.
\newblock Contactgrasp: Functional multi-finger grasp synthesis from contact.
\newblock In \emph{2019 IEEE/RSJ International Conference on Intelligent Robots and Systems (IROS)}, pp.\  2386--2393. IEEE, 2019.

\bibitem[Celik et~al.(2024)Celik, Taranovic, and Neumann]{celik2024acquiring}
Onur Celik, Aleksandar Taranovic, and Gerhard Neumann.
\newblock Acquiring diverse skills using curriculum reinforcement learning with mixture of experts.
\newblock \emph{arXiv preprint arXiv:2403.06966}, 2024.

\bibitem[Chao et~al.(2021)Chao, Yang, Xiang, Molchanov, Handa, Tremblay, Narang, Van~Wyk, Iqbal, Birchfield, et~al.]{dexycb}
Yu-Wei Chao, Wei Yang, Yu~Xiang, Pavlo Molchanov, Ankur Handa, Jonathan Tremblay, Yashraj~S Narang, Karl Van~Wyk, Umar Iqbal, Stan Birchfield, et~al.
\newblock Dexycb: A benchmark for capturing hand grasping of objects.
\newblock In \emph{Proceedings of the IEEE/CVF Conference on Computer Vision and Pattern Recognition}, 2021.

\bibitem[Cheng et~al.(2023)Cheng, Dong, Cai, and Sun]{cheng2023multi}
Guangran Cheng, Lu~Dong, Wenzhe Cai, and Changyin Sun.
\newblock Multi-task reinforcement learning with attention-based mixture of experts.
\newblock \emph{IEEE Robotics and Automation Letters}, 8\penalty0 (6):\penalty0 3812--3819, 2023.

\bibitem[Clevert(2015)]{clevert2015fast}
Djork-Arn{\'e} Clevert.
\newblock Fast and accurate deep network learning by exponential linear units (elus).
\newblock \emph{arXiv preprint arXiv:1511.07289}, 2015.

\bibitem[Davchev et~al.(2022)Davchev, Luck, Burke, Meier, Schaal, and Ramamoorthy]{davchev2022residual}
Todor Davchev, Kevin~Sebastian Luck, Michael Burke, Franziska Meier, Stefan Schaal, and Subramanian Ramamoorthy.
\newblock Residual learning from demonstration: Adapting dmps for contact-rich manipulation.
\newblock \emph{IEEE Robotics and Automation Letters}, 7\penalty0 (2):\penalty0 4488--4495, 2022.

\bibitem[Doya et~al.(2002)Doya, Samejima, Katagiri, and Kawato]{doya2002multiple}
Kenji Doya, Kazuyuki Samejima, Ken-ichi Katagiri, and Mitsuo Kawato.
\newblock Multiple model-based reinforcement learning.
\newblock \emph{Neural computation}, 14\penalty0 (6):\penalty0 1347--1369, 2002.

\bibitem[Fedus et~al.(2022)Fedus, Zoph, and Shazeer]{fedus2022switch}
William Fedus, Barret Zoph, and Noam Shazeer.
\newblock Switch transformers: Scaling to trillion parameter models with simple and efficient sparsity.
\newblock \emph{Journal of Machine Learning Research}, 23\penalty0 (120):\penalty0 1--39, 2022.

\bibitem[Garcia-Hernando et~al.(2020)Garcia-Hernando, Johns, and Kim]{garcia2020physics}
Guillermo Garcia-Hernando, Edward Johns, and Tae-Kyun Kim.
\newblock Physics-based dexterous manipulations with estimated hand poses and residual reinforcement learning.
\newblock In \emph{2020 IEEE/RSJ International Conference on Intelligent Robots and Systems (IROS)}, pp.\  9561--9568. IEEE, 2020.

\bibitem[Hang et~al.(2024)Hang, Lin, Zhu, Li, Wu, Ma, and Sun]{hang2024dexfuncgrasp}
Jinglue Hang, Xiangbo Lin, Tianqiang Zhu, Xuanheng Li, Rina Wu, Xiaohong Ma, and Yi~Sun.
\newblock Dexfuncgrasp: A robotic dexterous functional grasp dataset constructed from a cost-effective real-simulation annotation system.
\newblock In \emph{Proceedings of the AAAI Conference on Artificial Intelligence}, volume~38, pp.\  10306--10313, 2024.

\bibitem[Jacobs et~al.(1991)Jacobs, Jordan, Nowlan, and Hinton]{jacobs1991adaptive}
Robert~A Jacobs, Michael~I Jordan, Steven~J Nowlan, and Geoffrey~E Hinton.
\newblock Adaptive mixtures of local experts.
\newblock \emph{Neural computation}, 3\penalty0 (1):\penalty0 79--87, 1991.

\bibitem[Jia et~al.(2022)Jia, Li, Ling, Liu, Wu, and Su]{gsl}
Zhiwei Jia, Xuanlin Li, Zhan Ling, Shuang Liu, Yiran Wu, and Hao Su.
\newblock Improving policy optimization with generalist-specialist learning.
\newblock In \emph{International Conference on Machine Learning}, 2022.

\bibitem[Jiang et~al.(2024{\natexlab{a}})Jiang, Sablayrolles, Roux, Mensch, Savary, Bamford, Chaplot, Casas, Hanna, Bressand, et~al.]{jiang2024mixtral}
Albert~Q Jiang, Alexandre Sablayrolles, Antoine Roux, Arthur Mensch, Blanche Savary, Chris Bamford, Devendra~Singh Chaplot, Diego de~las Casas, Emma~Bou Hanna, Florian Bressand, et~al.
\newblock Mixtral of experts.
\newblock \emph{arXiv preprint arXiv:2401.04088}, 2024{\natexlab{a}}.

\bibitem[Jiang et~al.(2021)Jiang, Liu, Wang, and Wang]{jiang2021hand}
Hanwen Jiang, Shaowei Liu, Jiashun Wang, and Xiaolong Wang.
\newblock Hand-object contact consistency reasoning for human grasps generation.
\newblock In \emph{Proceedings of the IEEE/CVF international conference on computer vision}, pp.\  11107--11116, 2021.

\bibitem[Jiang et~al.(2024{\natexlab{b}})Jiang, Wang, Zhang, Wu, and Fei-Fei]{jiang2024transic}
Yunfan Jiang, Chen Wang, Ruohan Zhang, Jiajun Wu, and Li~Fei-Fei.
\newblock Transic: Sim-to-real policy transfer by learning from online correction.
\newblock \emph{arXiv preprint arXiv:2405.10315}, 2024{\natexlab{b}}.

\bibitem[Johannink et~al.(2019)Johannink, Bahl, Nair, Luo, Kumar, Loskyll, Ojea, Solowjow, and Levine]{johannink2019residual}
Tobias Johannink, Shikhar Bahl, Ashvin Nair, Jianlan Luo, Avinash Kumar, Matthias Loskyll, Juan~Aparicio Ojea, Eugen Solowjow, and Sergey Levine.
\newblock Residual reinforcement learning for robot control.
\newblock In \emph{2019 international conference on robotics and automation (ICRA)}, pp.\  6023--6029. IEEE, 2019.

\bibitem[Jordan \& Jacobs(1994)Jordan and Jacobs]{jordan1994hierarchical}
Michael~I Jordan and Robert~A Jacobs.
\newblock Hierarchical mixtures of experts and the em algorithm.
\newblock \emph{Neural computation}, 6\penalty0 (2):\penalty0 181--214, 1994.

\bibitem[Kappassov et~al.(2015)Kappassov, Corrales, and Perdereau]{tactile-dex}
Zhanat Kappassov, Juan-Antonio Corrales, and V{\'e}ronique Perdereau.
\newblock Tactile sensing in dexterous robot hands.
\newblock \emph{Robotics and Autonomous Systems}, 2015.

\bibitem[Lloyd(1982)]{kmeans}
Stuart Lloyd.
\newblock Least squares quantization in pcm.
\newblock \emph{IEEE transactions on information theory}, 1982.

\bibitem[Makoviychuk et~al.(2021)Makoviychuk, Wawrzyniak, Guo, Lu, Storey, Macklin, Hoeller, Rudin, Allshire, Handa, et~al.]{makoviychuk2021isaac}
Viktor Makoviychuk, Lukasz Wawrzyniak, Yunrong Guo, Michelle Lu, Kier Storey, Miles Macklin, David Hoeller, Nikita Rudin, Arthur Allshire, Ankur Handa, et~al.
\newblock Isaac gym: High performance gpu-based physics simulation for robot learning.
\newblock \emph{arXiv preprint arXiv:2108.10470}, 2021.

\bibitem[Mandikal \& Grauman(2022)Mandikal and Grauman]{mandikal2022dexvip}
Priyanka Mandikal and Kristen Grauman.
\newblock Dexvip: Learning dexterous grasping with human hand pose priors from video.
\newblock In \emph{Conference on Robot Learning}, pp.\  651--661. PMLR, 2022.

\bibitem[McKinzie et~al.(2024)McKinzie, Gan, Fauconnier, Dodge, Zhang, Dufter, Shah, Du, Peng, Weers, et~al.]{mckinzie2024mm1}
Brandon McKinzie, Zhe Gan, Jean-Philippe Fauconnier, Sam Dodge, Bowen Zhang, Philipp Dufter, Dhruti Shah, Xianzhi Du, Futang Peng, Floris Weers, et~al.
\newblock Mm1: Methods, analysis \& insights from multimodal llm pre-training.
\newblock \emph{arXiv preprint arXiv:2403.09611}, 2024.

\bibitem[Peng et~al.(2019)Peng, Chang, Zhang, Abbeel, and Levine]{peng2019mcp}
Xue~Bin Peng, Michael Chang, Grace Zhang, Pieter Abbeel, and Sergey Levine.
\newblock Mcp: Learning composable hierarchical control with multiplicative compositional policies.
\newblock \emph{Advances in neural information processing systems}, 32, 2019.

\bibitem[Pons et~al.(1999)Pons, Ceres, and Pfeiffer]{multifingered-review}
Jose~L Pons, R~Ceres, and Friedrich Pfeiffer.
\newblock Multifingered dextrous robotics hand design and control: a review.
\newblock \emph{Robotica}, 1999.

\bibitem[Qi et~al.(2017)Qi, Su, Mo, and Guibas]{pointnet}
Charles~R Qi, Hao Su, Kaichun Mo, and Leonidas~J Guibas.
\newblock Pointnet: Deep learning on point sets for 3d classification and segmentation.
\newblock In \emph{Proceedings of the IEEE conference on computer vision and pattern recognition}, 2017.

\bibitem[Qin et~al.(2022{\natexlab{a}})Qin, Huang, Yin, Su, and Wang]{dexpoint}
Yuzhe Qin, Binghao Huang, Zhao-Heng Yin, Hao Su, and Xiaolong Wang.
\newblock Dexpoint: Generalizable point cloud reinforcement learning for sim-to-real dexterous manipulation.
\newblock \emph{Conference on Robot Learning (CoRL)}, 2022{\natexlab{a}}.

\bibitem[Qin et~al.(2022{\natexlab{b}})Qin, Wu, Liu, Jiang, Yang, Fu, and Wang]{qin2022dexmv}
Yuzhe Qin, Yueh-Hua Wu, Shaowei Liu, Hanwen Jiang, Ruihan Yang, Yang Fu, and Xiaolong Wang.
\newblock Dexmv: Imitation learning for dexterous manipulation from human videos.
\newblock In \emph{European Conference on Computer Vision}, pp.\  570--587. Springer, 2022{\natexlab{b}}.

\bibitem[Rajeswaran et~al.(2017)Rajeswaran, Kumar, Gupta, Vezzani, Schulman, Todorov, and Levine]{rajeswaran2017learning}
Aravind Rajeswaran, Vikash Kumar, Abhishek Gupta, Giulia Vezzani, John Schulman, Emanuel Todorov, and Sergey Levine.
\newblock Learning complex dexterous manipulation with deep reinforcement learning and demonstrations.
\newblock \emph{arXiv preprint arXiv:1709.10087}, 2017.

\bibitem[Rana et~al.(2020)Rana, Talbot, Dasagi, Milford, and S{\"u}nderhauf]{rana2020residual}
Krishan Rana, Ben Talbot, Vibhavari Dasagi, Michael Milford, and Niko S{\"u}nderhauf.
\newblock Residual reactive navigation: Combining classical and learned navigation strategies for deployment in unknown environments.
\newblock In \emph{2020 IEEE International Conference on Robotics and Automation (ICRA)}, pp.\  11493--11499. IEEE, 2020.

\bibitem[Ross et~al.(2011)Ross, Gordon, and Bagnell]{dagger}
St{\'e}phane Ross, Geoffrey Gordon, and Drew Bagnell.
\newblock A reduction of imitation learning and structured prediction to no-regret online learning.
\newblock In \emph{Proceedings of the fourteenth international conference on artificial intelligence and statistics}, 2011.

\bibitem[Schoettler et~al.(2020)Schoettler, Nair, Luo, Bahl, Ojea, Solowjow, and Levine]{schoettler2020deep}
Gerrit Schoettler, Ashvin Nair, Jianlan Luo, Shikhar Bahl, Juan~Aparicio Ojea, Eugen Solowjow, and Sergey Levine.
\newblock Deep reinforcement learning for industrial insertion tasks with visual inputs and natural rewards.
\newblock In \emph{2020 IEEE/RSJ International Conference on Intelligent Robots and Systems (IROS)}, pp.\  5548--5555. IEEE, 2020.

\bibitem[Schulman et~al.(2017)Schulman, Wolski, Dhariwal, Radford, and Klimov]{schulman2017proximal}
John Schulman, Filip Wolski, Prafulla Dhariwal, Alec Radford, and Oleg Klimov.
\newblock Proximal policy optimization algorithms.
\newblock \emph{arXiv preprint arXiv:1707.06347}, 2017.

\bibitem[Shao et~al.(2020)Shao, Ferreira, Jorda, Nambiar, Luo, Solowjow, Ojea, Khatib, and Bohg]{shao2020unigrasp}
Lin Shao, Fabio Ferreira, Mikael Jorda, Varun Nambiar, Jianlan Luo, Eugen Solowjow, Juan~Aparicio Ojea, Oussama Khatib, and Jeannette Bohg.
\newblock Unigrasp: Learning a unified model to grasp with multifingered robotic hands.
\newblock \emph{IEEE Robotics and Automation Letters}, 5\penalty0 (2):\penalty0 2286--2293, 2020.

\bibitem[Shaw et~al.(2023)Shaw, Agarwal, and Pathak]{leaphand}
Kenneth Shaw, Ananye Agarwal, and Deepak Pathak.
\newblock Leap hand: Low-cost, efficient, and anthropomorphic hand for robot learning.
\newblock \emph{arXiv preprint arXiv:2309.06440}, 2023.

\bibitem[Shazeer et~al.(2017)Shazeer, Mirhoseini, Maziarz, Davis, Le, Hinton, and Dean]{shazeer2017outrageously}
Noam Shazeer, Azalia Mirhoseini, Krzysztof Maziarz, Andy Davis, Quoc Le, Geoffrey Hinton, and Jeff Dean.
\newblock Outrageously large neural networks: The sparsely-gated mixture-of-experts layer.
\newblock \emph{arXiv preprint arXiv:1701.06538}, 2017.

\bibitem[Silver et~al.(2018)Silver, Allen, Tenenbaum, and Kaelbling]{silver2018residual}
Tom Silver, Kelsey Allen, Josh Tenenbaum, and Leslie Kaelbling.
\newblock Residual policy learning.
\newblock \emph{arXiv preprint arXiv:1812.06298}, 2018.

\bibitem[Vincent(2011)]{score-matching-denoising}
Pascal Vincent.
\newblock A connection between score matching and denoising autoencoders.
\newblock \emph{Neural computation}, 2011.

\bibitem[Wan et~al.(2023)Wan, Geng, Liu, Shan, Yang, Yi, and Wang]{wan2023unidexgrasp++}
Weikang Wan, Haoran Geng, Yun Liu, Zikang Shan, Yaodong Yang, Li~Yi, and He~Wang.
\newblock Unidexgrasp++: Improving dexterous grasping policy learning via geometry-aware curriculum and iterative generalist-specialist learning.
\newblock In \emph{Proceedings of the IEEE/CVF International Conference on Computer Vision}, pp.\  3891--3902, 2023.

\bibitem[Wang et~al.(2023)Wang, Zhang, Chen, Xu, Li, Liu, and Wang]{wang2023dexgraspnet}
Ruicheng Wang, Jialiang Zhang, Jiayi Chen, Yinzhen Xu, Puhao Li, Tengyu Liu, and He~Wang.
\newblock Dexgraspnet: A large-scale robotic dexterous grasp dataset for general objects based on simulation.
\newblock In \emph{2023 IEEE International Conference on Robotics and Automation (ICRA)}, pp.\  11359--11366. IEEE, 2023.

\bibitem[Wei et~al.(2023)Wei, Wang, and Wang]{wei2023generalized}
Wei Wei, Peng Wang, and Sizhe Wang.
\newblock Generalized anthropomorphic functional grasping with minimal demonstrations.
\newblock \emph{arXiv preprint arXiv:2303.17808}, 2023.

\bibitem[Wu et~al.(2022)Wu, Guo, and Liu]{wu2022learning}
Albert Wu, Michelle Guo, and C~Karen Liu.
\newblock Learning diverse and physically feasible dexterous grasps with generative model and bilevel optimization.
\newblock \emph{arXiv preprint arXiv:2207.00195}, 2022.

\bibitem[Wu et~al.(2024{\natexlab{a}})Wu, Gan, Wu, Cheng, Yang, Zhu, and Dong]{wu2024unidexfpm}
Tianhao Wu, Yunchong Gan, Mingdong Wu, Jingbo Cheng, Yaodong Yang, Yixin Zhu, and Hao Dong.
\newblock Unidexfpm: Universal dexterous functional pre-grasp manipulation via diffusion policy.
\newblock \emph{arXiv preprint arXiv:2403.12421}, 2024{\natexlab{a}}.

\bibitem[Wu et~al.(2024{\natexlab{b}})Wu, Wu, Zhang, Gan, and Dong]{wu2024learning}
Tianhao Wu, Mingdong Wu, Jiyao Zhang, Yunchong Gan, and Hao Dong.
\newblock Learning score-based grasping primitive for human-assisting dexterous grasping.
\newblock \emph{Advances in Neural Information Processing Systems}, 36, 2024{\natexlab{b}}.

\bibitem[Xu et~al.(2023)Xu, Wan, Zhang, Liu, Shan, Shen, Wang, Geng, Weng, Chen, et~al.]{xu2023unidexgrasp}
Yinzhen Xu, Weikang Wan, Jialiang Zhang, Haoran Liu, Zikang Shan, Hao Shen, Ruicheng Wang, Haoran Geng, Yijia Weng, Jiayi Chen, et~al.
\newblock Unidexgrasp: Universal robotic dexterous grasping via learning diverse proposal generation and goal-conditioned policy.
\newblock In \emph{Proceedings of the IEEE/CVF Conference on Computer Vision and Pattern Recognition}, pp.\  4737--4746, 2023.

\bibitem[Yu et~al.(2020)Yu, Kumar, Gupta, Levine, Hausman, and Finn]{gradient-surgery}
Tianhe Yu, Saurabh Kumar, Abhishek Gupta, Sergey Levine, Karol Hausman, and Chelsea Finn.
\newblock Gradient surgery for multi-task learning.
\newblock \emph{Advances in Neural Information Processing Systems}, 2020.

\bibitem[Zhang et~al.(2023)Zhang, Wang, Sun, Wu, Zhu, and Tomizuka]{zhang2023efficient}
Xiang Zhang, Changhao Wang, Lingfeng Sun, Zheng Wu, Xinghao Zhu, and Masayoshi Tomizuka.
\newblock Efficient sim-to-real transfer of contact-rich manipulation skills with online admittance residual learning.
\newblock In \emph{Conference on Robot Learning}, pp.\  1621--1639. PMLR, 2023.

\end{thebibliography}
\bibliographystyle{iclr2025_conference}

\newpage
\appendix
% \newpage
\section{Appendix}
\subsection{Simulation Setup}
\label{appendix:setup}
%\red{object task set, detailed reward func, parallel env num, ...}

We conduct all our experiments in IsaacGym~\citep{makoviychuk2021isaac}, a GPU-accelerated platform for physics simulation and reinforcement learning. 
Each environment features a table that is 60 cm tall, with an object initialized 10 cm above the tabletop, which then falls onto it. 
The shadow hand is initialized 20 cm above the desktop. 
The task is to grasp the object and lift its center to 20 cm above the center of the tabletop.

The dataset is split into one training set and two test sets. 
The training set contains 3,200 object instances. The test sets include 141 instances of unseen objects within seen categories from the training set and 100 instances of unseen objects in unseen categories.
For state-based policies, we use PPO~\citep{schulman2017proximal} for training. 
For vision-based policies, we distill the state-based expert policy into a vision-based policy using DAgger~\citep{dagger}.
Each geometry-unaware policy is trained with 4,096 environments in parallel for 5,000 iterations. 
% Training on a single NVIDIA RTX 4090 GPU requires approximately 20 to 30 minutes.
The hyper policy is trained with 11,000 environments in parallel for 20,000 iterations for every training stage.
% , taking about 11 hours.
The vision-based policy is trained with 11,000 environments in parallel for 8000 iterations. %on an A800 GPU.
% , taking about 16.5 hours.

\textbf{Reward Function for Base Policy} \ We use a modified goal-conditioned reward function to train geometry-unaware base policies. The reward function is defined as:
$$r = r^{pose} + r^{task}$$
 $X_{joint}$ denotes the joint positions.  The $r^{pose}$ is defined as follows:
 $$r^{pose} = -0.05* \|\bm{q}-X_{joint}\|_1$$
% Details about $r^{pose}$ are in section \ref{sec:blindgrasp}. 

$r^{task}$ is defined as follows:
$$r^{task} = r^{reach} + r^{lift} + r^{move} + r^{bonus}$$

The $r^{reach}$ encourages the hand to reach the object, as it penalizes the distance between the object and different parts of the hand. Here, $X_{obj}$ and $X_{hand}$ denote the position of the object and the hand, and $X_{finger}$ denotes positions of all the fingers. The $r^{reach} $ is defined as follows:
    $$r^{reach} = -1.0*\|X_{obj}-X_{hand}\|_2 - 0.5*\sum\|X_{obj} - X_{finger}\|_2$$

The $r^{lift}$ encourages the hand to lift the object. It gives a positive reward when this condition can be satisfied: $f_1 = \bm{1}\left(\sum\|X_{obj} - X_{finger}\|_2\leq 0.6\right) + \bm{1}\left(\|X_{obj}-X_{hand}\|_2 \leq 0.12\right)$. $a_z$ is the scaled force applied to the hand root along the z-axis. The $r^{lift}$ is defined as follows:
\begin{equation*}
    r^{lift} = 
    \begin{cases}
        0.1+0.1*a_z & \text{if } f_1 = 2 \\
   0 & \text{otherwise}
    \end{cases}
\end{equation*}

The $r^{move}$ encourages the hand to move the object to the target position. $X_{target}$ denotes the target position. It gives a positive reward when this condition is satisfied: $f_2 = \bm{1}\left(\sum\|X_{obj} - X_{finger}\|_2\leq 0.6\right) + \bm{1}\left(\|X_{obj}-X_{hand}\|_2 \leq 0.12\right) + \bm{1}\left(\|\bm{q}-X_{joint}\|_1\leq 6\right) $. The $r^{move}$ is defined as follows:
\begin{equation*}
    r^{move} = 
    \begin{cases}
        0.9 - 2\|X_{obj}-X_{target}\|_2 & \text{if } f_2 = 3 \\
        0 & \text{otherwise}
    \end{cases}
\end{equation*}

The $r^{bonus}$ gives an extra reward when the object is close to the target position. We denote $\|X_{obj}-X_{target}\|_2$ as $d_{obj}$. The $r^{bonus}$ is defined as follows:
\begin{equation*}
    r^{bonus} = 
    \begin{cases}
        \frac{1}{1+10* d_{obj}} & \text{if } d_{obj} \leq 0.05 \\
        0 & \text{otherwise}
    \end{cases}
\end{equation*}

\textbf{Reward Function for Hyper Policy} \ At the first training stage for a hyper policy, we use the goal-conditioned reward function exactly the same as the one proposed in UniDexGrasp\citep{xu2023unidexgrasp}.

At the second training stage for a hyper policy, we use a loosened reward function defined as follows:
$$ r = r^{lift} + r^{move} + r^{bonus}$$
The definitions of $r^{lift}$ and $r^{bonus}$ are the same as those mentioned above. The $r^{move}$ has loosened its condition. It is defined as follows:
\begin{equation*}
    r^{move} = 
    \begin{cases}
        0.9 - 2\|X_{obj}-X_{target}\|_2 & \text{if } f_1 = 2 \\
        0 & \text{otherwise}
    \end{cases}
\end{equation*}

\subsection{Training Details}
\label{appendix:train-detail}
%\red{network arch, hyperparams, device and training time, ...}

\textbf{Network Architecture} \ We use a MLP architecture which consists of 4 layers (1024, 1024, 512, 512) for base policies and the hyper policy. For the vision-based policy, we use a simplified PointNet \citep{pointnet} encoder to represent the object point cloud and apply MLPs with the same hidden layer sizes for the actor and the critic. We use ELU~\citep{clevert2015fast} as the activation function. 

\textbf{Training Device and Training Time} \ All the state-based policies are trained on on a single NVIDIA RTX 4090 GPU. Training a base policy takes about 20 minutes, while training a hyper-policy takes about 11 hours. 
For the vision-based policy, we train on a single A800 GPU, taking about 16 hours.

The hyperparameters of PPO and DAgger are described in Table \ref{tab:ppo-param} and Table \ref{tab:dagger-param}.

%\newpage

\renewcommand{\arraystretch}{1}
\begin{table}[htbp]
  \caption{Hyperparameters of PPO.}
  \label{tab:ppo-param}
  \centering
  \begin{tabular}{ccc}
    \toprule
    Name & Symbol & Value \\
    \midrule
    Episode length & \--\-- & 200 \\
    Num. envs (base policy) & \--\-- & 4096 \\
    Num. envs (hyper-policy)& \--\-- & 11000 \\
    Parallel rollout steps per iteration & \--\-- & 8 \\
    Training epochs per iteration & \--\-- & 5 \\
    Num. minibatches per epoch & \--\-- & 4 \\
    Optimizer & \--\-- & Adam \\
    Clip gradient norm & \--\-- &1.0 \\
    %Desired KL  &  \--\-- &0.016 \\
    Initial noise std. & \--\-- &0.8 \\
    Clip observations & \--\-- & 5.0 \\
    Clip actions & \--\-- & 1.0 \\ 
    %Entropy coeff.& \--\-- & 0.0 \\ 
    Learning rate & $\eta$ & 3e-4 \\
    Discount factor & $\gamma$ & 0.96 \\
    GAE lambda & $\lambda$ & 0.95 \\
    Clip range & $\epsilon$ & 0.2 \\
    
    \bottomrule
  \end{tabular}
\end{table}

\begin{table}[htbp]
  \caption{Hyperparameters of DAgger.}
  \label{tab:dagger-param}
  \centering
  \begin{tabular}{ccc}
    \toprule
    Name & Symbol & Value \\
    \midrule
    Episode length & \--\-- & 200 \\
    Num. envs & \--\-- & 11000 \\
    Parallel rollout steps per iteration & \--\-- & 1 \\
    Training epochs per iteration & \--\-- & 5 \\
    Num. minibatches per epoch & \--\-- & 4 \\
    Optimizer & \--\-- & Adam \\
    Clip observations & \--\-- & 5.0 \\
    Clip actions & \--\-- & 1.0 \\ 
    Learning rate & $\eta$ & 3e-4 \\
    Clip range & $\epsilon$ & 0.2 \\
    \bottomrule
  \end{tabular}
\end{table}

%\section{Additional Results}
%\label{appendix:result}

\end{document}